\documentclass[sigconf]{acmart}

\pdfoutput=1
\usepackage{hyperref}

\usepackage{booktabs} 

\usepackage[T1]{fontenc}
\usepackage[latin9]{inputenc}
\usepackage{geometry}
\usepackage{array}
\usepackage{float}
\usepackage{wrapfig}
\usepackage{booktabs}
\usepackage{graphicx}
\usepackage{setspace}

\usepackage{algorithm}
\usepackage{algpseudocode}



\renewcommand\footnotetextcopyrightpermission[1]{} 
\setcopyright{none}

\begin{document}
\title[Identifying Significant Predictive Bias in Classifiers]{Identifying Significant Predictive Bias in Classifiers}
\subtitle{June 2017}

\author{Zhe Zhang} 
\affiliation{%
  \institution{Carnegie Mellon University}   
  \streetaddress{5000 Forbes Ave}   
  \city{Pittsburgh}    
  \state{PA}    
  \postcode{15213} 
}
\email{zhezhang@cmu.edu}

\author{Daniel Neill} 
\affiliation{%
  \institution{Carnegie Mellon University}   
  \streetaddress{5000 Forbes Ave}   
  \city{Pittsburgh}    
  \state{PA}    
  \postcode{15213} 
}
\email{neill@cs.cmu.edu}

\renewcommand{\shortauthors}{Zhe Zhang and Daniel Neill}

\begin{abstract}
We present a novel subset scan method to detect if a probabilistic
binary classifier has statistically significant bias --- over or under
predicting the risk --- for some subgroup, and identify the characteristics
of this subgroup. This form of model checking and goodness-of-fit
test provides a way to interpretably detect the presence of classifier
bias or regions of poor classifier fit. This allows consideration
of not just subgroups of a priori interest or small dimensions, but
the space of all possible subgroups of features. To address the difficulty
of considering these exponentially many possible subgroups, we use
subset scan and parametric bootstrap-based methods. Extending this
method, we can penalize the complexity of the detected subgroup and
also identify subgroups with high classification errors. We demonstrate
these methods and find interesting results on the COMPAS crime recidivism
and credit delinquency data.
\end{abstract}

\maketitle


\section{Introduction}

Increasingly, data-driven tools like probabilistic classifiers are
used for decision support and risk assessment in various sectors:
criminal justice, public policy, health, banking, online platforms
\cite{angwin2016machine,goel2016precinct,miller2015algorithms,starr2014sentencing}.
To evaluate the usage of such methods, we usually focus on overall
predictive performance. However, recent academic and popular writing
has also emphasized the importance of potential biases or discrimination
in these predictions. Earlier this year, ProPublica conducted a widely
discussed analysis \cite{angwin2016machine} of the COMPAS recidivism
risk prediction algorithm, arguing that the predictions, controlling
for actual risk, were more likely to mistakenly predict black defendants
as high-risk of reoffending. 

Bias in data-driven classifiers could have several possible sources
and forms of bias. We focus on the source of bias from classification
techniques that could be insufficiently flexible to predict well for
subgroups in the data, due to optimizing for overall performance or
model mis-specification. In this paper, we focus on the \emph{predictive
bias }in probabilistic classifiers or risk prediction that result
from that source. As a simplified example --- we define predictive
bias in detail in Section 2 --- consider a subgroup $S$, with outcomes
$Y\in\{0,1\}$ and a classifier's predictions $0<\hat{p}_{S}<1$ for
that subgroup's outcomes; over-estimation predictive bias is considered:
\[
\mathbb{P}(Y=1|1_{\{S\}})<\hat{p}_{S}
\]
here $1_{\{S\}}$ is an indicator function for membership in a subgroup
$S$, and vice-versa for under-estimation. Predictive bias is different
from predictive fairness, which emphasizes comparable predictions
between subgroups of a priori interest, like race or gender, while
predictive bias emphasizes comparable predictions for a subgroup and
its observations.

In this paper, we (1) define a measure of predictive bias based on
how a subgroup's observed outcome odds are different from the predicted
odds, and (2) operationalize this definition into a \emph{bias scan}
method to detect and identify which subgroup(s) have statistically
significant predictive bias, given a classifier's predictions $\hat{p}$.
Further, we discuss briefly extending this method to penalize subgroup
complexity or detect subgroups with higher than expected classification
errors, and present novel case study results from the bias scan on
crime recidivism and loan delinquency predictions.

Existing literature on predictive bias focus on sets of subgroups
$\mathcal{S}$ defined by one dimension of a priori interest, such
as race, gender, or income. However, some important subgroups may
not be described so simply or be considered a priori. ProPublica's
COMPAS analysis \cite{angwin2016machine} and follow-up analyses
\cite{chouldechova2017fair,skeem2016risk} focus on predictive bias
for subgroups $\mathcal{S}=\{white,black\}$. In our analysis of COMPAS,
we do not detect a significant predictive bias along racial lines,
but instead identify bias in a more subtle multi-dimensional subgroup:
females who initially committed misdemeanors (rather than felonies),
for half of the COMPAS risk groups, have their recidivism risk significantly
over-estimated.

Assessing bias in all the exponentially many subgroups is a difficult
task both computationally and statistically. First, exhaustively evaluating
all subgroups for predictive bias quickly becomes computationally
infeasible. From a dataset with $M$ features, with each feature having
$|X_{m}|$ discretized values, we define a subgroup as any $M$-dimension
Cartesian set product, between subsets of feature-values from each
feature --- excluding the empty set. With this axis-aligned criteria,
we only consider subgroups that are interpretable, rather than a collection
of dataset rows. There are then $\prod_{m=1}^{M}\left(2^{|X_{m}|}-1\right)$
unique subgroups; consider a dataset with only 4 discretized features
(e.g. age, income, ethnicity, location), each with arity $|X_{m}|=5$,
then there are $\approx10^{6}$ possible subgroups to consider (and
$\approx10^{7}$ if there were a fifth feature).

A second difficulty is estimating statistical significance of a detected
subgroup. It is trivial to identify \emph{some }measure of predictive
bias --- any subgroup where fraction of observed outcomes does not
equal the predicted proportion. The relevant question is instead,
can we identify a subgroup(s) that has significantly more predictive
bias than would be expected from an unbiased classifier?

To address these difficulties, this work develops a novel extension
of fast subset scan anomaly detection methods \cite{neill2012fast,neill2013fast,kumar2012fast,speakman2016penalized}.
This enables our bias scan method to approximately identify the most
statistically biased subgroup in linear time (rather than exponential).
We then use parametric bootstrapping \cite{efron1994introduction}
to adjust for multiple testing and estimate the statistical significance
of the detected subgroup. A distinguishing mechanism of our method
is the ability to statistically consider all possible subgroups, expanding
the search space beyond just interaction effects, and in essence,
enabling the collective consideration of groups of weak, but related
signals.

\paragraph{Related Literature}

Topically, the problem of assessing bias in data-driven decision making
covers areas including predictive bias, problems in the original training
data, disparate impacts of predictions, and adjusting predictions
to ensure fairness \cite{adler2016auditing,dwork2012fairness,feldman2015certifying,romei2014multidisciplinary}
--- in areas like criminal justice \cite{angwin2016machine,chouldechova2017fair,flores2016false,skeem2016risk},
but also other various sectors \cite{house2016big,o2016weapons,miller2015algorithms,starr2014sentencing}.
We focus here on predictive bias. Existing literature on predictive
bias has focused just on subgroups of a priori interest, such as race
or gender \cite{angwin2016machine,chouldechova2017fair,flores2016false,skeem2016risk}.
We contribute by providing a more general method that can detect and
characterize such bias, or poor classifier fit, in the larger space
of all possible subgroups, without a priori specification.

Methodologically, our method is comparable to other methods that analyze
the residuals between a classifier's predictions $\hat{p}$ and observed
outcomes $Y$. This includes a range of long-standing literature,
including model checking, goodness-of-fit methods, and visualization
of residuals. The identification of patterns in residuals is an early
key lesson when teaching regression, first using one-dimensional visualization.
A common more rigorous extension is to use interpretable predictive
methods to characterize patterns in residuals. Examples of methods
include linear models using quadratic or interaction terms. These
models cannot collectively consider groups of signals or interactions
though, unless specified ex ante, e.g. group lasso \cite{yuan2006model}.
More flexible assessment of patterns in residuals, comparing residual
sum-of-squares between linear models and non-linear methods like random
forests, can be formally used via a generalized F-test-style test
and parametric bootstrapping, as Shah and Buhlmann \cite{RSSB:RSSB12234}
show for regression. They aim to detect the general presence of poor
fit for model selection, but do not characterize where this bias is.
Tree-based methods with top-down optimization may split subgroups
of interest and need distinguishing significance of leaves.

\section{Bias Subset Scan Methodology}

We extend methodology from the anomaly detection literature, specifically
the use of fast, expectation-based subset scans \cite{neill2012fast,neill2013fast,kumar2012fast,speakman2016penalized}.
This methodology is able to identify or approximate the most anomalous
subgroup of feature space in linear time, amongst the exponentially
many possible ones, enabling tractable subgroup analysis. The general
form of these methods are:
\[
S^{*}=FSS(\mathcal{D},\mathcal{E},F_{score})
\]
where $S^{*}$ is the detected most anomalous subgroup, $FSS$ is
one of several subset scan algorithms for different problem settings,
$\mathcal{D}$ is a dataset with outcomes $Y$ and discretized features
$\mathcal{X}$, $\mathcal{E}$ are a set of expectations or ``normal''
values for $Y$, and $F_{score}$ is an expectation-based scoring
statistic that measures the amount of anomalousness between subgroup
observations and their expectations. For this to be tractable, the
$F_{score}$ statistic must satisfy Linear Time Subset Scanning (LTSS,
\cite{neill2012fast}) or Additive Linear Time Subset Scanning (ALTSS,
\cite{speakman2016penalized}) properties --- which prove that feature-values
for one feature can be optimally ordered to reduce the number of subgroups
to consider.

In the bias scan method, we develop a novel extension specifically
of the Multi-Dimensional Subset Scan (MDSS) method \cite{neill2013fast},
described by Algorithm 1. We contribute (1) a new subgroup scoring
statistic, $score_{bias}$, that measures the bias in a given subgroup,
and prove it satisfies the ALTSS property; and (2) the application
of parametric bootstrapping for the subset scanning setting to estimate
statistical significance of detections.

First, we define the statistical measure of predictive bias function,
$score_{bias}(S)$. It is a likelihood ratio score and a function
of a given subgroup $S$. The null hypothesis is that the given prediction's
odds are correct for all subgroups in $\mathcal{D}$: $H_{0}:odds(y_{i})=\frac{\hat{p}_{i}}{1-\hat{p}_{i}}\ \forall i\in\mathcal{D}$.
The alternative hypothesis assumes some constant multiplicative bias
in the odds for some given subgroup $S$:
\[
H_{1}:\ odds(y_{i})=q\frac{\hat{p}_{i}}{1-\hat{p}_{i}},\ \text{where}\ q>1\ \forall i\in S\ \mbox{and}\ q=1\ \forall i\notin S.
\]
In the classification setting, each observation's likelihood is Bernoulli
distributed and assumed independent. This results in the following
scoring function for a subgroup $S$\footnote{Following common practice in the scan statistics literature, we maximize
the free parameter $q$ to identify the most likely alternative hypothesis,
which also maximizes the score. The resulting score is influenced
by both the number of observations in the subgroup $s$ and most likely
value of $q$.}:
\begin{align*}
score_{bias}(S)= & \max_{q}\log\prod_{i\in S}\frac{Bernoulli(\frac{q\hat{p}_{i}}{1-\hat{p}_{i}+q\hat{p}_{i}})}{Bernoulli(\hat{p}_{i})}\\
= & \max_{q}\log(q)\sum_{i\in S}y_{i}-\sum_{i\in S}\log(1-\hat{p}_{i}+q\hat{p}_{i}).
\end{align*}
Our bias scan is thus represented as: $S^{*}=FSS(\mathcal{D},\mathcal{E},F_{score})=MDSS(\mathcal{D},\hat{p},score_{bias})$.

Second, to determine the statistical p-value that the given classifier/model
has a biased subgroup, we use parametric bootstrap simulated predictions
under the null of a correctly specified classifier, as also used by
\cite{RSSB:RSSB12234} in their residual prediction tests. As extensions,
with little added computation, we also introduce penalties for subgroup
complexity, with an increasing penalty for each feature, based on
the size of each feature's subset of feature-values, but with no penalty
if the feature includes 1 or all feature-values. This encourages a
reduced dimension of the detected subgroup, and can be used in an
``elbow-curve'' style heuristic between bias score and complexity.
Also, to detect subgroups with higher than expected classification
errors, we can adjust the bias scan based on the concept that a subgroup's
predictions $\hat{p}$ also predicts the expected classification rate
of a subgroup.
\begin{algorithm}
\footnotesize{
\begin{algorithmic}
\caption{Pseudocode for Multi-dimensional Subset Scan}
\label{alg:mdss}
\State{Initialize $best\_score, i, cur\_subgroup$;}
\State{$cur\_data\_subset=Data|_{cur\_subgroup}$}

\Repeat
\State{1. Randomly order the given $m$ features to scan from 1 to $M$}

\For{$j=1\text{ to } M$}
\State{1. $cur\_data\_subset = Data|_{cur\_subgroup_{-j}}$}
\State{(relax the subgroup definition to include all values of feature $j$)}
\State{2. $cur\_subgroup = MDSS(cur\_dataset)$}
\State{(Use MDSS on $cur\_dataset$ to identify the exact highest scoring subset of values of feature $j$, given $cur\_subgroup_{-j}$)}
\State{3. $cur\_data\_subset = Data|_{cur\_subgroup}$}
\State{4. $best\_score = score_{bias}(cur\_dataset)$}
\EndFor
\State{2. Check end condition, else loop through features in random order again, $i = i + 1$}
\Until{$best\_score$ has not changed between $i$ and $i-1$}

\end{algorithmic}
}
\end{algorithm}

\section{Demonstrative Bias Scan Results}

\begin{figure*}
\begin{centering}
\includegraphics[height=1.4in]{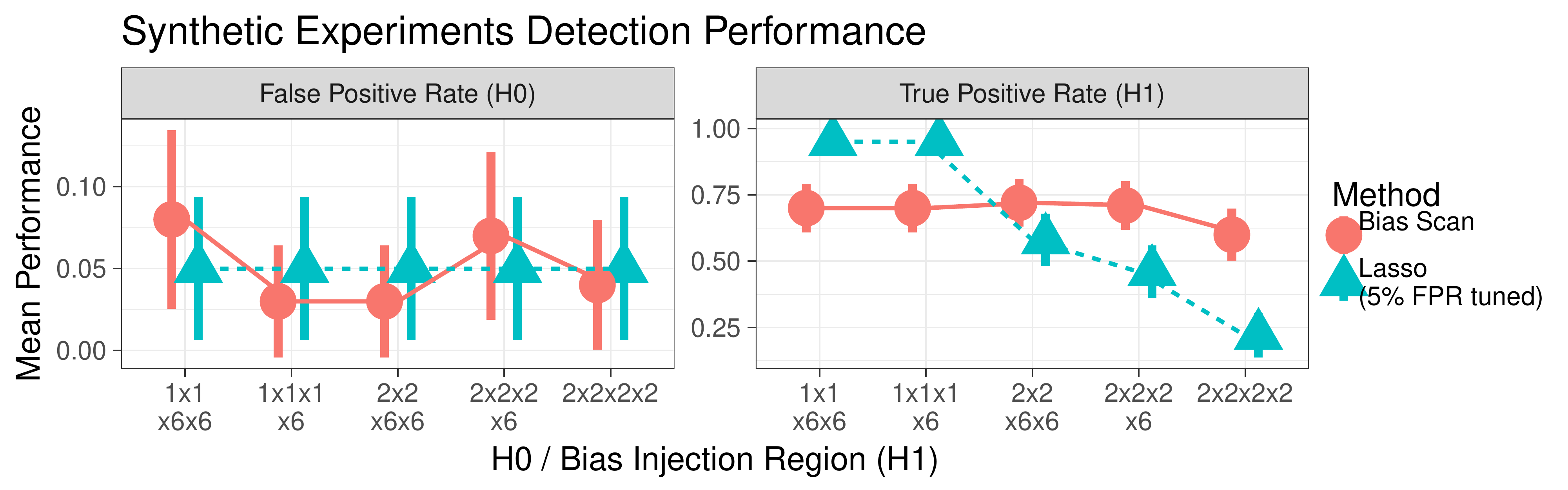}
\par\end{centering}
\begin{centering}
\includegraphics[height=1.4in]{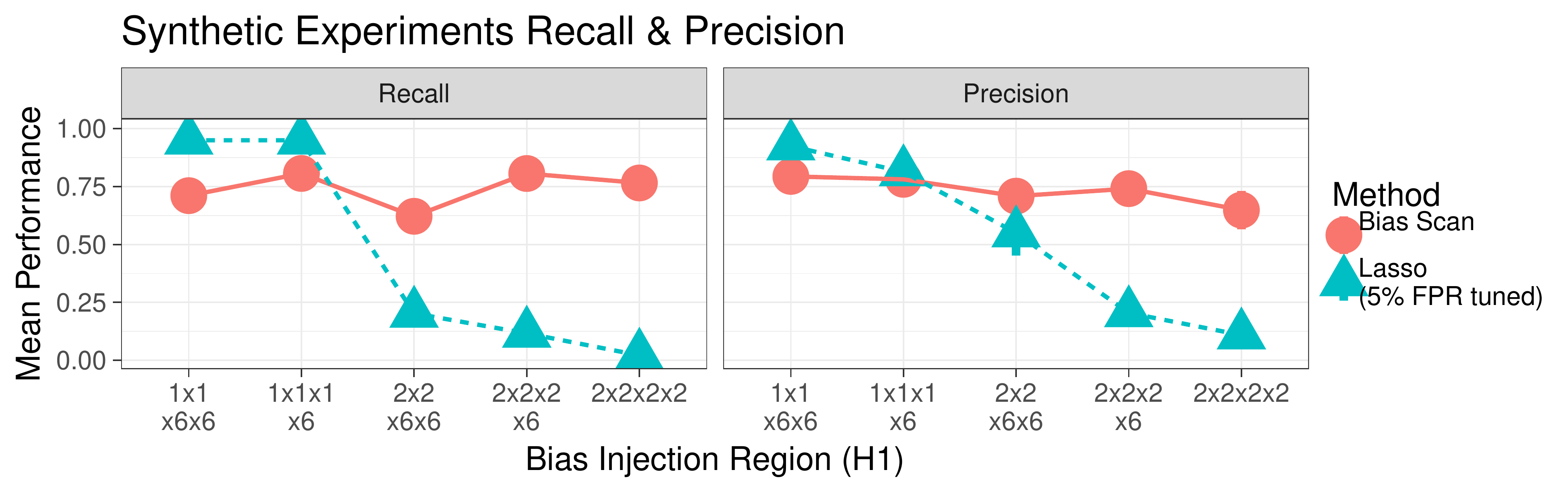}
\par\end{centering}
\centering{}\caption{The detection performance of a lasso analysis of residuals versus
the bias scan, including confidence intervals. The first two columns
represent one 2 or 3-way interaction with bias; the other three represent
injected bias across multiple 2, 3, or 4-way interactions.}
\label{fig:synthetic}
\end{figure*}

In synthetic experiments, we compare the bias scan method to both
a lasso and stepwise regression analysis of residuals\footnote{We calibrate the lasso analysis of residuals to identify the lasso
penalty hyper-parameter that has a 5\% false positive rate (on data
with no bias injections). This is to match the chosen 5\% false positive
rate of the bias-scan, and is visualized in the top-left of Figure
\ref{fig:synthetic}. Though we only inject one subgroup with bias
and the lasso can detect multiple subgroups simultaneously, the bias
scan could also be used repeatedly to detect multiple subgroups.}. The detection performance results are shown in Figure \ref{fig:synthetic},
without the stepwise regression however, because it always has worse
detection performance than the lasso regression analysis. In the experiment,
we generate data with 4 categorical features, each with arity 6, and
data is evenly distributed and generate Bernoulli outcomes\footnote{Each feature-value has random coefficient values in an additive log-odds
model (e.g. a logistic regression model) to generate probabilities
for each observation and draw Bernoulli outcomes.}. The variation in the experiment is in the injected predictive bias:
we inject additional log-odds bias of size 1.5 in one, or several,
interaction effects of 2, 3, or 4 dimensions. To demonstrate the power
of grouping weak signals, we only affect 100 observations, but range
from concentrating them all in one specific interaction, or spread
them across several related interactions\footnote{The total number of observations changes depending on the size of
the injected region, to ensure that data is evenly distributed across
all feature-values.}.

We find that the lasso analysis on residuals, which considers the
space of all 2, 3, and 4-way interactions and uses the cross-validation
optimal penalty\footnote{Using the ``1SE'' penalty term has worse detection performance.},
has a better rate of non-zero interaction coefficients (top-right
of Figure \ref{fig:synthetic}) when the injected bias is concentrated
in one 2-way or 3-way interaction. However, when the injected bias
is spread across four 2-way interactions, eight 3-way interactions,
or sixteen 4-way interactions, lasso detection rate is below the bias
scan detection rate, e.g., approximately 25\% compared to 60\%. A
similar story, where the lasso has more difficulty when bias is spread
across related interactions, occurs looking at the recall and detection
precision of the biased observations/tensor cells, shown in the bottom
half of Figure \ref{fig:synthetic}. For example, when the bias is
injected in eight (8) 3-way interactions (\emph{``2x2x2x6}''): lasso
has an average recall/precision of 35\%/45\% compared to the average
bias scan recall/precision of 75\%/80\%. This highlights the potential
improvement by considering subgroups rather than interactions, grouping
weak, related signals together. If we used the bias scan to detect
a subgroup to use for an additional logistic regression model term,
this also has improved out-of-sample prediction performance over a
lasso logistic regression with 2, 3, and 4-way interaction effects.

\paragraph{Recidivism Prediction Case Study}

As a case study for identifying biases in classification, we apply
our bias scan method to the COMPAS crime recidivism risk prediction
dataset provided by ProPublica $(n=6172)$. This dataset includes
age, race, gender, number of prior offenses, and crime severity (felony
vs misdemeanor) for each individual, along with a binary gold standard
label (of reoffending in a 2-year time period) and the classification
prediction made by the COMPAS algorithm (categorized risk groups 1,
2, ..., 10). We find notable biases by the COMPAS prediction that
we have not seen noted elsewhere. We assume the provided decile scores
adequately represent all the private information that COMPAS uses.
We initialize by fitting an unpenalized logistic regression based
on categorized decile scores. Using bias scan, we find that the COMPAS
decile scores clearly have predictive bias on subgroups defined by
their counts of priors. Defendants with >5 priors are significantly
under-estimated by the COMPAS deciles (mean predicted rate of 0.60
in the subgroup, observed rate of 0.72, $n=1215$), while those with
0 priors are significantly over-estimated (mean predicted rate of
0.38, observed rate of 0.29, $n=2085$).

\begin{figure}
\begin{centering}
\includegraphics[height=1.65in]{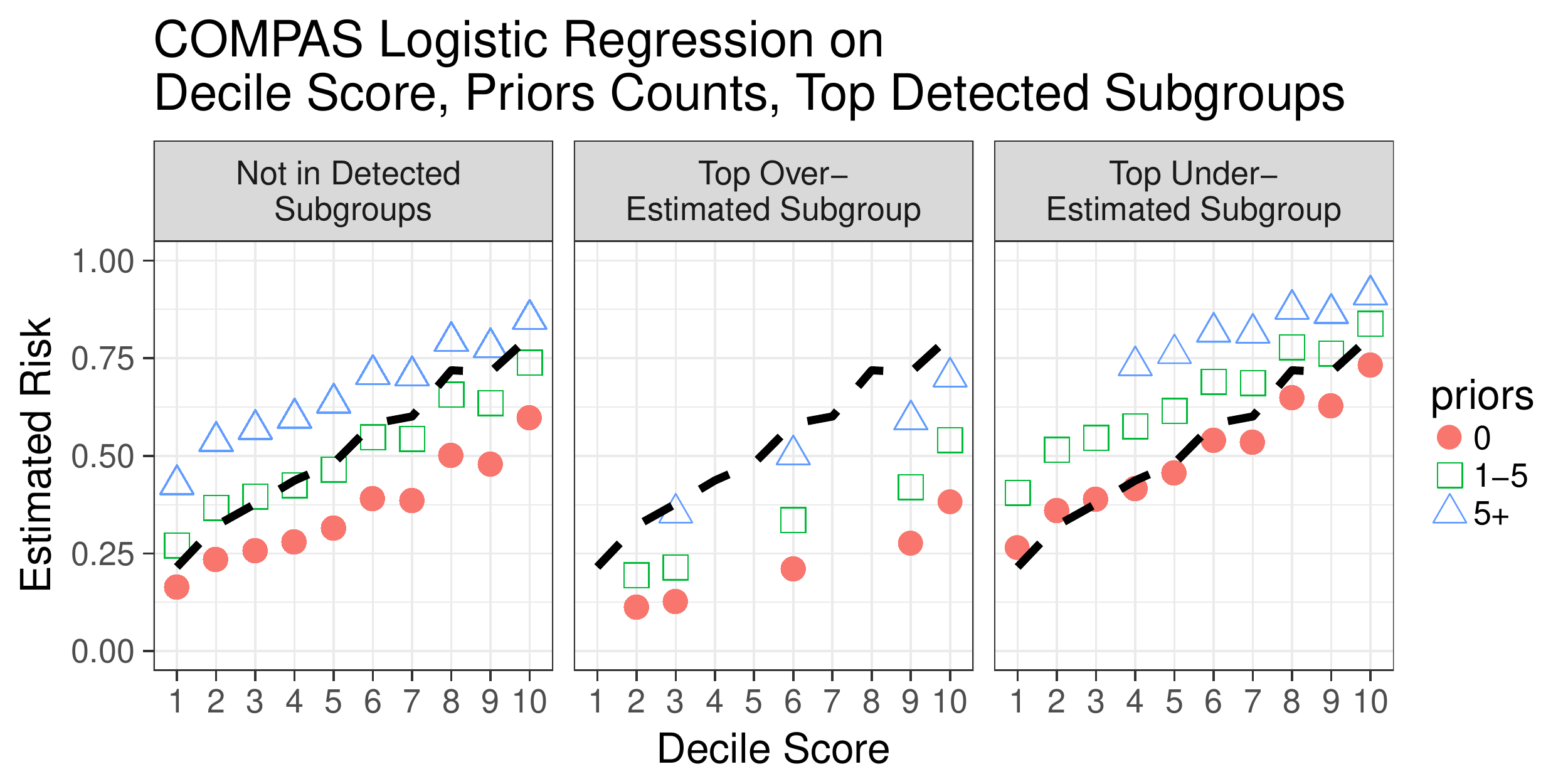}
\par\end{centering}
\centering{}\caption{Predictions from the baseline COMPAS decile model are a black dashed
line, the subgroup adjusted predictions are the colored points.}
\label{fig:compas}
\end{figure}

Using this initial finding, we refit the model to account for both
decile score and discretized prior counts. Applying the bias scan
again on the predictions of this improved classifier, we again identify
two significant subgroups of classifier bias. Young (< 25 years) males
are under-estimated (regardless of race or initial crime type) ($p<0.005$);
with an observed recidivism rate of 0.60 and a predicted rate of 0.50
($n=1101$). Additionally, females, whose initial crimes were misdemeanors,
and have COMPAS decile scores $\in\{2,3,6,9,10\}$ are over-estimated
($p=0.035$); with an observed recidivism rate of 0.21 and a predicted
rate of 0.38 ($n=202$). In Figure \ref{fig:compas}, we compare the
original COMPAS decile model (black dashed line), with the logistic
model that accounts for the four detected subgroups.

The over- and under-estimated subgroups involve 2 features and 3 features,
respectively. They were identified by penalizing the complexity of
the detected subgroup. The unpenalized detected subgroups had scores
of $(28.6,17.0)$ respectively and involved 4 and 5 features respectively.
When we add some penalty terms for complexity, we slightly reduce
the scores to $(24.5,16.0)$ respectively, both of which are still
significant at a 5\% FPR, but reduces the number of involved features
to 2 and 3, respectively. For the young males, it removed involvement
of race and COMPAS decile score; for the females with misdemeanors,
it removed involvement of features on priors and race.

\noindent 

\paragraph{Other Datasets of Interest}

Expanding our analysis, we identify predictive bias from the use of
various classifiers (e.g., lasso regression on all 2-way interactions,
tree-based classifiers) applied to various datasets (credit risk,
stop-and-frisk weapon carrying prediction {[}based on the stop-and-frisk
data-driven model proposed by \cite{goel2016precinct}{]}, income
prediction, COMPAS, breast cancer prediction, and diabetes prediction).
For each type of classifier, we detect significant subgroups of predictive
bias in some of those datasets. Furthermore, when we hold out half
of the dataset, we find the significant detected subgroups also have
the same directional bias in the held-out data, though the magnitude
of the bias was smaller, as expected.

As an example, we discuss the credit delinquency prediction dataset,
i.e. ``Give Me Some Credit'' dataset provided by Kaggle. In this
dataset, using the cross-validation optimal lasso regression on all
the discretized features, the top identified over-estimated subgroup
is defined by those users in the top half of utilization in the data
(>15\% credit limit utilization) and who have at least 1 occurrence
of being within each of 30-59, 60-89, and 90+ days late (i.e., on
a 3 separate payments); ($p<0.01$, observed rate of 2-year delinquency
of 0.79, predicted rate of 0.90). There are $n=825$ such accounts
in the dataset, about 1.7\% of the total dataset. For comparison,
the mean rate of delinquency in the entire total dataset is 15\%.
In this same data, we detect a high error subgroup, with both a predicted
and observed rate of 61\%, but with much more classification errors
than expected due to over-confidence by the classifier on both low
and high predicted-risk consumers.

To understand the potential impact of this predictive bias, consider
if this data were used to rank customers by their delinquency risk.
470 of the 496 top 1\% riskiest consumers belong to the over-estimated
subgroup. If the observations in that over-estimated subgroup were
adjusted by a constant multiplication to their predicted odds, then
only 286 consumers from that subgroup would then be ranked in the
top 1\%.


\bibliographystyle{ACM-Reference-Format}
\bibliography{zhang_neill_fatml_17} 

\end{document}